\PassOptionsToPackage{table}{xcolor}
\PassOptionsToPackage{numbers,square}{natbib}
\documentclass[11pt,a4paper,logo]{googledeepmind}
\setleftlogo[180pt]{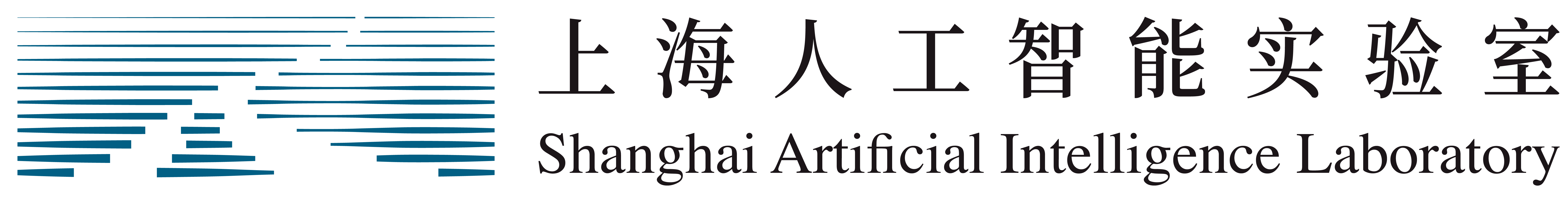}
\setrightlogo{}
\usepackage[numbers,square]{natbib}

\usepackage{amsmath,amsfonts,bm}

\def\eqref#1{equation~\ref{#1}}

\def\1{\bm{1}}

\DeclareMathAlphabet{\mathsfit}{\encodingdefault}{\sfdefault}{m}{sl}
\SetMathAlphabet{\mathsfit}{bold}{\encodingdefault}{\sfdefault}{bx}{n}

\usepackage{enumitem}
\usepackage[utf8]{inputenc}
\usepackage[T1]{fontenc}
\usepackage{microtype}
\usepackage{soul}
\usepackage{hyperref}
\usepackage{url}
\usepackage{fontawesome5}

\usepackage{amsmath,amssymb,amsfonts}
\usepackage{nicefrac}

\usepackage{graphicx}
\usepackage{subcaption}
\usepackage{booktabs}
\usepackage[table]{xcolor}   
\definecolor{methodhighlight}{HTML}{FFF3DD}
\usepackage{multirow}
\usepackage{makecell}
\usepackage{tabularx}
\usepackage{array}
\usepackage{wrapfig}
\usepackage{placeins}

\newcolumntype{C}{>{\centering\arraybackslash}X}
\newcolumntype{L}{>{\raggedright\arraybackslash}X}
\newcolumntype{R}{>{\raggedleft\arraybackslash}X}

\usepackage{pifont}

\newcommand{\sname}{MLExpert\xspace}
\usepackage{xspace}

\title{\sname: Towards a Generalist Agentic Model for Autonomous Machine Learning Engineering}
\codeurl{https://github.com/InternScience/MLExpert}
\modelurl{https://huggingface.co/InternScience/MLExpert}

\author[1 2 $\spadesuit$]{Shangheng Du}
\author[1 2 $\spadesuit$]{Jinxin Shi}
\author[2]{Shuaiyu Zhang}
\author[2]{Yiming Wu}
\author[2\ ]{Junpeng Zhao}
\author[2\ ]{Shiyang Feng}
\author[2\ ]{Xin Li}
\author[1 2\ ]{Liang He}
\author[2 $\clubsuit\diamondsuit$]{Xiangchao Yan}
\author[2 $\clubsuit$]{Bo Zhang}
\affil[1]{East China Normal University}
\affil[2]{Shanghai Artificial Intelligence Laboratory}
\correspondingauthor{%
$\spadesuit$: Co-first Authors \quad $\diamondsuit$: Project Lead \quad $\clubsuit$: Corresponding Authors\\
Please send correspondence regarding this report to \href{mailto:yanxiangchao@pjlab.org.cn}{yanxiangchao@pjlab.org.cn} and \href{mailto:zhangbo@pjlab.org.cn}{zhangbo@pjlab.org.cn}.%
}

\hypersetup{colorlinks=true,linkcolor=blue,citecolor=blue,filecolor=black,urlcolor=blue}

\begin{document}

\sloppy

\begin{abstract}
Large language models (LLMs) are expanding from assisting scientific research and engineering to autonomously developing and improving AI systems, advancing AI for AI (AI4AI) as a pathway toward recursive self-improvement (RSI). 
Workflow-driven machine learning engineering (MLE) systems provide a practical path toward this pursuit through complex orchestration that manages long-horizon experimentation, while recent MLE systems adopt model-directed interaction, allowing models to autonomously organize experiments and adapt optimization strategies based on execution feedback. However, feedback from each experiment reflects several preceding decisions, and a long MLE run contains many such costly experiments. Training must therefore select useful supervision from long interaction trajectories and learn from intermediate outcomes as they become available. To address these challenges, we propose a post-training method for autonomous long-horizon MLE optimization and apply it to train \sname{}, a 35B Mixture-of-Experts agentic model. We design the Kaggle-MLE Factory to construct executable tasks and an interaction harness for model-directed experimentation. Hierarchical agentic SFT transfers general tool-use and coding capabilities to MLE through outcome-guided supervision. Windowed Interaction RL (WIRL) treats the multi-turn decisions between successive experiment results as an execution window and assigns it a shared outcome-based reward. Asynchronous collection and policy correction support policy updates before the trajectory ends. On MLE-Bench Lite, \sname{} improves the average medal rate of the base model from 37.9\% to 72.7\%, achieving performance comparable to larger frontier models such as Kimi-K3 and GLM-5.3. Further experiments show that \sname{} improves performance on NatureBench Lite, agent scaffold optimization, and algorithm optimization, suggesting that the learned optimization capability generalizes to other tasks.
\end{abstract}

\maketitle
\section{Introduction}

The rapid advancement of reasoning, code generation, and tool interaction in large language models (LLMs) is enabling artificial intelligence (AI) to move beyond assisting scientific research and engineering toward autonomously developing and improving AI systems.
These efforts include model training, optimization of solver programs, and algorithm design, making AI for AI (AI4AI) a research direction toward recursive self-improvement (RSI).
In this context, machine learning engineering (MLE) provides a concrete setting for this goal, requiring long-horizon optimization through repeated execution, feedback analysis, and revision rather than one-shot generation.

\begin{figure}[thbp] \centering \vspace{-0.5em} \includegraphics[width=0.9\linewidth]{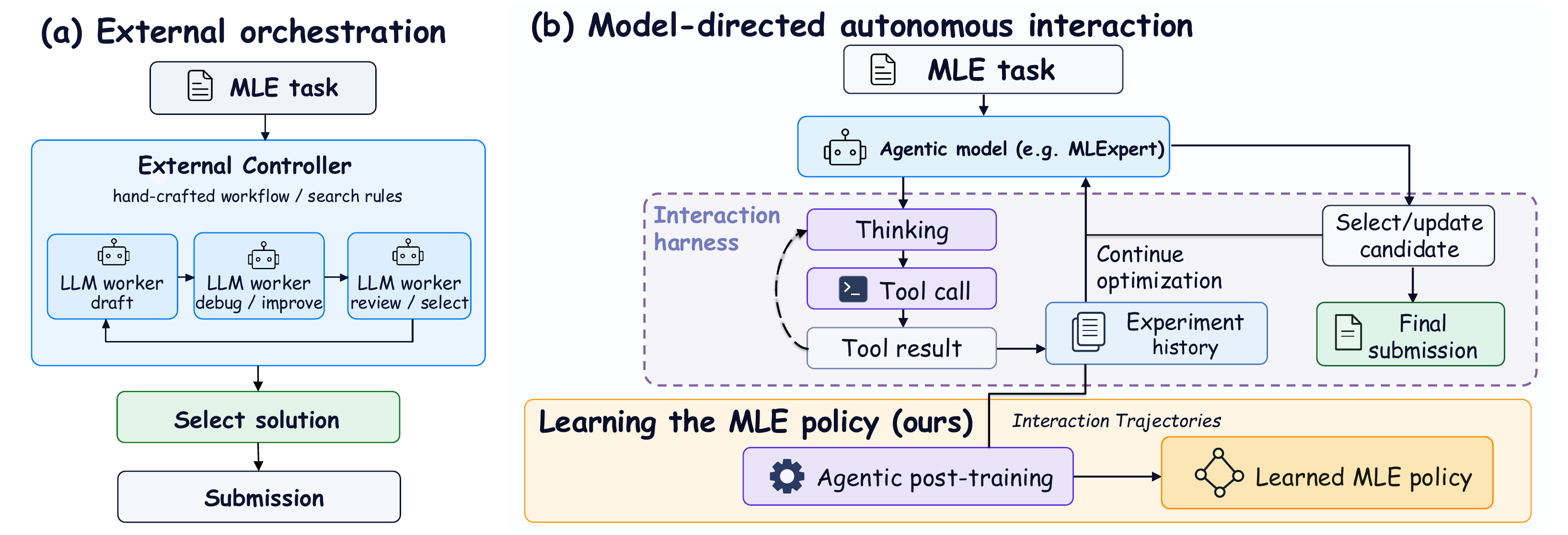} \vspace{-0.5em} \caption{Decision making in long-horizon MLE optimization. (a) External orchestration organizes optimization through predefined workflows and search rules. (b) In model-directed interaction, the model selects subsequent actions based on experimental states and execution feedback.} \label{intro_fig} \end{figure}

Most existing MLE systems~\citep{aide,mle-star,rdagent} extend code generation into iterative optimization through search, memory, or multi-agent collaboration.
As shown in Figure~\ref{intro_fig}(a), predefined workflows or search rules typically control experiment sequencing and stage transitions, while LLMs primarily generate or debug code at designated steps.
These systems rely heavily on manually designed control logic to organize long-horizon experimentation.

Recently, Arbor~\citep{jin2026arbor} and Argus~\citep{li2026argus} maintain persistent research states and allow models to design, continue, and revise experiments, placing greater demands on the model’s ability to plan experiments and sustain optimization over long interaction horizons.
ML-Agent~\citep{liu2025mlagent} combines multi-turn demonstrations with single-step RL, while Frontis-MA1~\citep{yang2026frontis} trains program evolution operators through SFT and RL and integrates them into an external evolutionary search framework.
These approaches primarily train models for single-step decisions or local operations within externally organized processes.
In contrast, we study how to train models to organize experiments and improve solutions over sustained interaction, using experience and execution feedback from real-world MLE tasks.
However, training this long-horizon optimization capability poses three challenges.
1) Open-ended MLE requires an interaction environment that preserves experimental states while leaving action selection to the model.
2) High-quality long-horizon supervision is difficult to obtain because MLE tasks are heterogeneous and interaction trajectories may mix useful exploration and failure recovery with degraded decisions.
3) Each costly experiment returns feedback on several preceding decisions before the full run ends; online RL must use this intermediate signal while handling policy mismatch during asynchronous collection.

To address these challenges, we propose a post-training method for autonomous MLE that combines a model-directed long-horizon interaction harness, hierarchical agentic SFT, and Windowed Interaction Reinforcement Learning (WIRL).
Based on this, we train \textbf{\sname{}}, a 35B Mixture-of-Experts (MoE) agentic model, to learn long-horizon optimization strategies.
As shown in Figure~\ref{intro_fig}(b), \sname{} uses sustained tool interaction within the interaction harness to autonomously organize experiments, revise solutions based on execution feedback, and maintain submission candidates.
We build the Kaggle-MLE Factory to convert historical Kaggle competitions into executable MLE tasks.
Hierarchical agentic SFT transfers general tool-use and coding capabilities to MLE optimization, with execution outcomes guiding the selection of trajectories and segments for supervision.
To use this intermediate feedback, WIRL treats each completed experiment as the boundary of an execution window containing decisions since the previous result. It trains those decisions with a shared outcome-based reward, while asynchronous collection and policy correction allow updates as trajectories continue.
On MLE-Bench Lite, \sname{} improves the average medal rate of the base model from 37.9\% to 72.7\%, achieving performance comparable to larger models such as Kimi-K3 and GLM-5.3.
\sname{} also improves performance on NatureBench Lite and transfers its optimization capabilities to agent scaffold and algorithm optimization tasks.
Our key contributions are:

\begin{itemize}[leftmargin=*]

\item We introduce \textbf{\sname{}}, a 35B MoE model for autonomous MLE, trained with hierarchical agentic SFT and WIRL. WIRL links execution feedback to multi-turn decisions within execution windows and combines local progress rewards with policy correction for asynchronous online learning.

\item We build the \textbf{Kaggle-MLE Factory} and an \textbf{Interaction Harness} to provide executable tasks and model-directed long-horizon interaction for training and evaluation.

\item Extensive experiments show that \sname{} improves the base model's average medal rate from 37.9\% to 72.7\% on MLE-Bench Lite, improves performance on NatureBench Lite, and generalizes to agent scaffold and algorithm optimization tasks.

\end{itemize}

\section{Preliminary}
\label{sec:mle}
\label{sec:agentic_mle}

We formulate MLE as the iterative optimization of executable ML solutions within a given time budget $B_e$.
An MLE task is represented as $e=(d_e,\mathcal{D}_e,m_e)$, where $d_e$ is the task description, $\mathcal{D}_e$ is the task data, and $m_e$ is the evaluation metric.
A candidate solution $y\in\mathcal{Y}_e$ represents a complete ML pipeline including data processing, feature engineering, model training, and prediction, where $\mathcal{Y}_e$ denotes the solution space.
The objective is to find a best-performing solution for a given task, $y_e^\star \in \arg\max_{y\in\mathcal{Y}_e} S_e(y)$,
where $S_e(y)$ denotes the task-specific score under metric $m_e$, normalized so that higher values indicate better performance.
Following MLE-bench~\citep{mle-bench}, the agent has access to the task description and public data, and improves solutions through code execution and self-constructed validation.
The final submission is independently evaluated using hidden test labels.

In Agentic MLE, an LLM policy $\pi_\theta$ autonomously organizes the optimization process.
Given a task $e$, the model constructs, executes, and improves candidate solutions through continuous tool interaction within the time budget $B_e$.
A trajectory of length $L$ is represented as $\tau=(o_0,a_0,o_1,a_1,\ldots,a_{L-1},o_L)$,
where $o_0$ is the initial task observation, $a_l$ is the model action at step $l$, and $o_{l+1}$ is the observation returned after executing $a_l$.
Each action may include model reasoning and one or more tool calls.
At step $l$, the model samples $a_l\sim\pi_\theta(\cdot\mid c_l)$, where $c_l$ consists of the task information and retained interaction history.

Experimental states, including code, execution records, and candidate solutions, are stored in an external workspace and exposed to the model through tool observations.
Rather than following a predefined workflow, the model decides when to analyze, revise, execute, or change the optimization direction based on the current state.
The model maintains a submission candidate throughout the interaction and selects its final solution $\hat{y}_\tau$ at the end of trajectory $\tau$ based on the experimental results and analysis available during interaction.
The objective of Agentic MLE is to maximize the expected quality of this selected solution
\begin{equation}
\label{eq:agentic_mle_objective}
J(\pi_\theta;e)
=
\mathbb{E}_{\tau\sim p_{\pi_\theta}(\cdot\mid e,B_e)}
\left[
S_e(\hat{y}_\tau)
\right],
\end{equation}
where $p_{\pi_\theta}$ denotes the trajectory distribution induced by the policy and the interaction environment.

\section{Task Construction and Long-Horizon Interaction}

\begin{figure}[t]
    \centering
    \includegraphics[width=\linewidth]{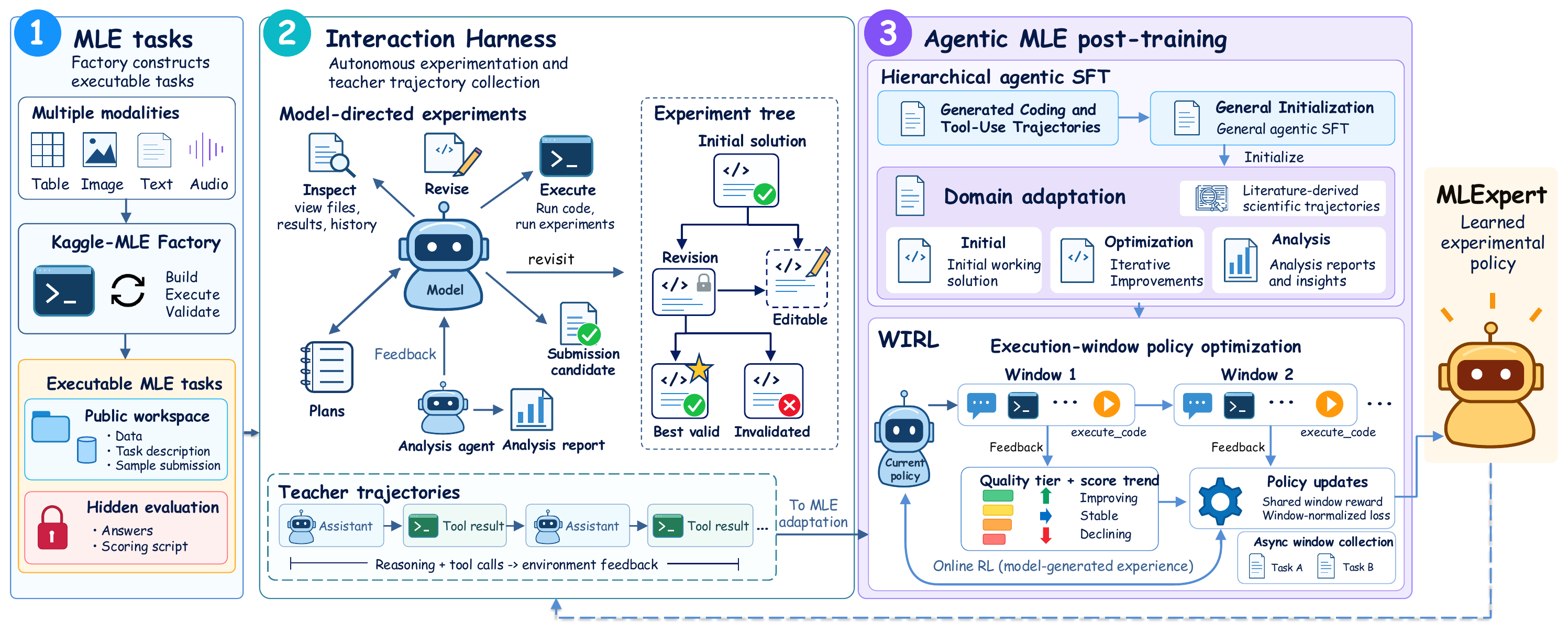}
    \caption{\textbf{Overview of \sname{} training for autonomous MLE.}
    The Kaggle-MLE Factory constructs executable tasks, and the interaction harness supports model-directed experimentation and trajectory collection.
    Hierarchical agentic SFT learns from demonstrations, while WIRL further optimizes the policy from execution feedback collected during online interaction.}
    \label{fig:method_overview}
\end{figure}

As shown in Figure~\ref{fig:method_overview}, we build the \textbf{Kaggle-MLE Factory} to provide executable tasks and the \textbf{Interaction Harness} to support model-directed long-horizon experimentation and state management, together enabling hierarchical agentic SFT and WIRL for post-training \sname{} in autonomous MLE.

\subsection{Kaggle-MLE Factory: Constructing Executable MLE Tasks}
\label{sec:kaggle_factory}

Long-horizon Agentic MLE training requires executable tasks covering diverse data modalities, objectives, and domains.
Historical Kaggle competitions provide a rich source of real-world MLE problems, but their heterogeneous data organization, submission formats, and evaluation protocols make manual task adaptation difficult to scale.
Therefore, we build the Kaggle-MLE Factory, which converts historical competitions into executable tasks with a public workspace, a hidden test split, and a unified evaluation interface.
The Factory uses a multi-agent pipeline for task construction, validation, and repair.
The discovery agent filters competitions by status, scale, and submission format.
After competition data, leaderboard information, and official descriptions are collected, the analyzer agent extracts data structures, fields, and modality characteristics to form a task specification.
The generator agent then prepares task data, constructs a held-out test split with hidden labels from the original labeled data, and implements the task-specific evaluator.
The task is executed on real data and iteratively repaired based on validator checks of data integrity, submission format, and evaluation behavior to produce valid executable tasks.
Standardized task packages support different MLE agent frameworks.
Final manual review excludes tasks with unresolved label or scoring issues before trajectory collection.
Appendix~\ref{app:kaggle_factory} details agent responsibilities and implementation.

\subsection{Interaction Harness for Long-Horizon MLE Optimization}
\label{sec:harness}

To support model-directed long-horizon MLE optimization, we design an interaction harness that follows ReAct-style loop~\citep{yao2022react}, leaving branch selection and optimization-path decisions to the model rather than an external search policy.
The harness stores experimental states in solution nodes, each containing candidate code, execution logs, and metadata. Parent-linked revisions form an experiment tree that the model can revisit and compare to decide whether to continue, backtrack, or change optimization direction.
The harness provides tools for solution analysis, code revision, experiment execution, history retrieval, and submission management.
The model launches candidate execution through \texttt{execute\_code} and updates its submission candidate through \texttt{update\_answer}.
The complete tool list and descriptions are provided in Appendix~\ref{app:harness_tools}.

As interaction history approaches the context limit, earlier messages are summarized while recent interactions are retained.
For complex analysis, \texttt{analyze} delegates task analysis, data exploration, and result diagnosis to an analysis sub-agent with independent context, returning only a summary to the main model.
Since overfitting or data leakage can inflate validation scores, the model can call \texttt{invalidate\_node} based on experimental evidence and analysis to exclude unreliable candidates.
With new evidence, the model can update its submission candidate while exploring alternatives.

\section{Agentic MLE Post-Training}
\label{sec:post_training}


\subsection{Hierarchical Agentic SFT}
\label{sec:hierarchical_sft}

Long-horizon MLE optimization requires general tool-use, coding skills, and domain-specific decision making.
We therefore use two-stage hierarchical agentic SFT.
In the first stage, coding and tool-use trajectories from diverse agentic tasks are used to learn basic tool-use and code revision capabilities.
The second stage adopts teacher trajectories from MLE domain and literature-derived scientific tasks to improve algorithm design, experimental analysis, and iterative optimization.

\noindent\textbf{Complementary MLE demonstrations.}
MLE adaptation data cover both task initialization and sustained optimization.
For trajectories whose first complete solution executes successfully and reaches a quality threshold, we extract the interaction prefix from task start through the corresponding execution feedback as an \textit{initial-solution trajectory}, typically covering task understanding, algorithm design, and code generation.
We also use \textit{long-horizon optimization trajectories} from complete interactions, covering solution construction, feedback analysis, iterative revision, and final submission selection.
In addition, analysis sub-agent trajectories are adopted to train the model for independent data analysis and result diagnosis.

\noindent\textbf{Outcome-guided supervision selection.}
Long-horizon MLE trajectories can contain effective exploration and failure recovery, but may also include misguided decisions or continued degradation after reaching a strong solution.
Therefore, we use execution results of submitted solutions to select high-quality trajectories and the segments used for supervision.
For trajectories that pass a quality threshold, we identify the first best-performing submission and measure the subsequent performance degradation.
When the post-peak degradation exceeds a threshold, we use the first best-performing submission as the anchor for retaining a trajectory prefix; otherwise, we use the last submission. The prefix ends at a complete tool-interaction boundary near the anchor.
This removes degraded tails while preserving useful exploration and failure recovery before the peak.

\noindent\textbf{Supervision samples and training objective.}
To preserve multi-turn context dependencies, each supervision sample combines existing interaction context with a continuous sequence of subsequent model actions and environment feedback from the selected MLE interactions.
When context compaction occurs, the retained history and summary form the conditioning context for subsequent samples.
Training loss is applied only to newly generated reasoning, text responses, and tool-call tokens, while tool feedback and inherited interaction history are used only as conditioning context and masked out from the loss.
Details of outcome-guided filtering, score normalization, and the SFT objective appear in Appendix~\ref{app:sft_details}.


\subsection{Windowed Interaction Reinforcement Learning}
\label{sec:wirl}

Building on SFT, we further optimize the long-horizon decision policy using execution feedback from the model's own exploration.
In MLE, solution quality is available only after training and evaluation, and each result depends on multiple rounds of analysis, code revision, and tool interaction.
We introduce Windowed Interaction RL (WIRL), which groups consecutive model decisions made between successive experiment results into an execution window $u_m$.
Each window is conditioned on the accumulated interaction history and associated with the outcome of the corresponding experiment launched through the \texttt{execute\_code} tool (Sec.~\ref{sec:harness}). The first window starts from the initial task state.
This allows policy updates to use feedback from completed experiments without waiting for the entire task trajectory to finish.

\noindent\textbf{Window rewards and shared optimization weights.}
MLE tasks use different evaluation metrics and score scales, so we use task-specific medal thresholds as a common reference for solution quality and compare each valid submission with the most recent valid submission in the trajectory to capture local optimization progress.
The reward of execution window $u_m$ is
\begin{equation}
\label{eq:window_reward}
R_m=
\begin{cases}
-1.0, & \text{execution error caused by model actions},\\
-0.6, & \text{no such error but no scorable submission},\\
-0.4+T_m, & \text{valid submission below the medal threshold},\\
1.0+T_m, & \text{bronze or silver medal},\\
2.0+T_m, & \text{gold medal}.
\end{cases}
\end{equation}

Here, $T_m$ denotes the local performance trend and is set to $0$ for the first valid submission.
For subsequent valid submissions, $T_m\in\{-0.2,0,0.2\}$ according to whether performance decreases, remains unchanged, or improves relative to the previous valid submission, accounting for the direction of the task metric.
Unlike standard GRPO~\citep{shao2024deepseekmath}, WIRL directly uses the verifiable execution reward as the shared optimization weight $\widehat A_m=R_m$, without group-relative normalization, and applies it to all main-model output tokens included in training within window $u_m$.

\noindent\textbf{Asynchronous collection and policy correction.}
Since experiment durations vary, WIRL asynchronously collects windows with trusted execution feedback to form training batches.
A long trajectory may span multiple policy updates, so its model calls may be generated by rollout policies that differ from the current training policy.
We therefore apply token-level policy correction using log-probabilities recorded during collection.
For output token $t$ from model call $j$ in window $u_m$, let $\ell_{\theta,mjt}$, $\ell^{\mathrm{old}}_{mjt}$, and $\ell^{\mathrm{roll}}_{mjt}$ denote its log-probabilities under the current training policy, the policy fixed before the update, and the rollout policy, respectively:
\begin{equation}
\label{eq:policy_correction}
q_{mjt}
=
\exp\!\left(
\ell^{\mathrm{old}}_{mjt}-\ell^{\mathrm{roll}}_{mjt}
\right),
\qquad
c_{mjt}
=
\operatorname{sg}\!\left(
q_{mjt}\mathbf{1}[q_{\min}\leq q_{mjt}\leq q_{\max}]
\right),
\end{equation}
where $\operatorname{sg}$ denotes stop-gradient, $\mathbf{1}[\cdot]$ is the indicator function, and $q_{\min}$ and $q_{\max}$ are the lower and upper correction bounds.
Tokens within $[q_{\min},q_{\max}]$ are weighted by $q_{mjt}$, while those outside the interval are excluded from the policy-gradient update.

\noindent\textbf{Policy Optimization Objective.}
Let $L^{\mathrm{PG}}_{mjt}$ denote the PPO-style token-level policy-gradient loss~\citep{schulman2017proximal} with shared window weight $\widehat A_m=R_m$.
Because execution windows vary in the number of model calls and output tokens, we normalize the loss by the number of valid output tokens within each window before averaging across windows.
Let $M_{mjt}$ denote the training mask for newly generated main-model output tokens and $N_m=\sum_{j,t}M_{mjt}$ the number of valid output tokens in window $u_m$.
For a batch $\mathcal{B}$ containing $B$ windows, we minimize
\begin{equation}
\label{eq:wirl_objective}
\mathcal{L}_{\mathcal{B}}(\theta)
=
\frac{1}{B}\sum_{m\in\mathcal{B}}
\frac{1}{\max(N_m,1)}
\sum_{j,t}M_{mjt}
\left[
c_{mjt}L^{\mathrm{PG}}_{mjt}
+\lambda_{\mathrm{KL}}K_{mjt}
-\lambda_H H_{mjt}
\right],
\end{equation}
where $K_{mjt}$ is the KL regularization term relative to a fixed SFT reference model, $H_{mjt}$ is the entropy of the current policy, and $\lambda_{\mathrm{KL}}$ and $\lambda_H$ are their coefficients.
The full policy loss definition and implementation details are provided in Appendix~\ref{app:wirl_details}.

\section{Experiments}

\subsection{Experimental Setup}
\label{sec:exp_setup}

\noindent\textbf{Benchmarks.}
We primarily evaluate our trained models on MLE-Bench Lite~\citep{mle-bench}, a 22-task subset of MLE-Bench consisting of historical Kaggle competitions for end-to-end machine learning engineering.
We report the percentage of tasks awarded each medal tier (Gold, Silver, and Bronze), the percentage receiving any medal (Any Medal), and the percentage exceeding the median human score (Med+), with Any Medal as the primary metric.
All five metrics are computed over the 22 tasks and reported as percentages.
We also evaluate on NatureBench~\citep{wang2026naturebench}, following the ten-task Lite subset used in Frontis-MA1~\citep{yang2026frontis}.

\noindent\textbf{Baselines.}
We compare \sname{} against two groups of models:
(1) \textit{same-scale models}, including Qwen3.6-35B-A3B~\citep{qwen36_35b_a3b}, Ornith-1.5-35B-A3B~\citep{ornith_1_5}, and Nex-N2.5-mini~\citep{nex_n25_mini};
and (2) \textit{larger frontier models}, including DeepSeek-V4-Pro Preview~\citep{xu2026deepseek}, MiniMax M3~\citep{minimax_m3}, GLM-5.3~\citep{zeng2026glm}, Kimi-K3~\citep{team2026kimi}, GPT-5.5~\citep{openai_gpt55}, and GPT-5.6 Sol~\citep{openai_gpt56_sol}.

\noindent\textbf{Implementation details.} Each MLE task is assigned a maximum of 3,000 interaction steps and a 12-hour runtime, using 20 vCPUs, 220 GB of RAM, and a single NVIDIA H200 GPU.
\sname{} is initialized from Qwen3.6-35B-A3B.
General SFT uses coding and tool-use trajectories from StepFun~\citep{huang2026step}.
For MLE SFT, we collect teacher trajectories from Kaggle-MLE Factory tasks, MLE-Dojo~\citep{qiang2026mledojo} tasks, and scientific tasks derived from Nature-family papers.
Any training tasks overlapping with MLE-Bench or NatureBench~\citep{wang2026naturebench} are excluded.
Data statistics, training hyperparameters, and implementation details are provided in Appendix~\ref{app:posttraining_details}.

\subsection{Main Results}
\label{sec:main_results}

\begin{table}[t]
\centering
\caption{\textbf{Main results on MLE-Bench Lite.}
We report performance (\%) under the unified \sname{} Harness and, where available, under general-purpose agentic coding harnesses. Bold and underline mark the best and second-best distinct values in each metric column, including ties. Detailed evaluation results are provided in Appendix~\ref{app:mle_results}.}
\label{tab:mle_main_results}
\begingroup
\setlength{\tabcolsep}{2.7pt}
\renewcommand{\arraystretch}{1.18}
\begin{tabularx}{\linewidth}{@{}lccccc>{\centering\arraybackslash}Xc@{}}
\toprule
\multirow{2}{*}[-0.3ex]{\textbf{Model}} & \multicolumn{5}{c}{\textbf{\sname{} Harness}}
& \multicolumn{2}{c}{\textbf{General-Purpose Harness}} \\
\cmidrule(lr){2-6}\cmidrule(lr){7-8}
& \textbf{Med+}
& \makecell[c]{\textbf{Gold}}
& \makecell[c]{\textbf{Silver}}
& \makecell[c]{\textbf{Bronze}}
& \textbf{Medal}
& \makecell[c]{\textbf{Harness}}
& \textbf{Medal} \\
\midrule

\addlinespace[0.15em]
\multicolumn{8}{@{}l}{\textcolor[HTML]{666666}{\textit{Same-scale baselines (35B total, $\sim$3B active)}}} \\[0.1em]
Qwen3.6-35B-A3B
& 62.1 & 22.7 & 7.6 & \underline{7.6} & 37.9
& Claude Code & 42.4 \\
Ornith-1.5-35B-A3B
& 57.6 & 22.7 & 12.1 & 4.5 & 39.4
& Claude Code & 47.0 \\
Nex-N2.5-mini
& 66.7 & 24.2 & 13.6 & \textbf{9.1} & 47.0
& Claude Code & 36.4 \\

\midrule
\addlinespace[0.15em]
\multicolumn{8}{@{}l}{\textcolor[HTML]{666666}{\textit{Larger frontier models}}} \\[0.1em]
DeepSeek-V4-Pro Preview
& 86.4 & 40.9 & 13.6 & \textbf{9.1} & 63.6
& -- & -- \\
MiniMax M3
& 81.8 & 54.5 & 13.6 & 0.0 & 68.2
& Codex & 54.5 \\
GLM-5.3
& \underline{90.9} & \underline{59.1} & 13.6 & 0.0 & \underline{72.7}
& Claude Code & \textbf{77.3} \\
Kimi-K3
& \textbf{95.5} & \textbf{63.6} & 9.1 & 0.0 & \underline{72.7}
& Claude Code & \underline{72.7} \\
GPT-5.5 xhigh
& 81.8 & 45.5 & \textbf{22.7} & 4.5 & \underline{72.7}
& Codex & 68.2 \\
GPT-5.6 Sol
& \underline{90.9} & \underline{59.1} & 13.6 & 4.5 & \textbf{77.3}
& Codex & \underline{72.7} \\

\midrule
\rowcolor{methodhighlight}
\textbf{\sname{} (Ours)}
& 84.8 & 50.0 & \underline{16.7} & 6.1 & \underline{72.7}
& Claude Code & 69.7 \\

\bottomrule
\end{tabularx}
\endgroup
\end{table}

\noindent\textbf{\sname{} achieves the highest medal rate among same-scale models and reaches the performance level of larger frontier models.}
As shown in Table~\ref{tab:mle_main_results}, under the unified \sname{} Harness, \sname{} raises the medal rate of its Qwen3.6-35B-A3B base model from 37.9\% to 72.7\%, achieving the highest rate among the same-scale models evaluated and performance comparable to GLM-5.3 and Kimi-K3.
These results show that long-horizon agentic post-training substantially improves the base model's MLE capability, bringing a 35B model to a performance level comparable to larger frontier models.

\noindent\textbf{\sname{} retains its advantage over the base model under a general-purpose coding harness.}
With Claude Code, \sname{} achieves a 69.7\% medal rate, compared with 42.4\% for its base model, and exceeds the strongest evaluated same-scale baseline by 22.7 percentage points.
The consistent advantage over the base model across both harnesses suggests that the improvement from post-training is not specific to the \sname{}-Harness and remains effective in a general-purpose agentic coding environment.

\begin{table}[t]
\centering
\caption{\textbf{Results on NatureBench Lite.} M denotes Match-SOTA ($g\geq0$), while S denotes Surpass-SOTA ($g>0.1$), which serves as a key metric for directly assessing agents' ability to surpass published scientific results.}
\label{tab:naturebench_results}
\begingroup
\setlength{\tabcolsep}{4pt}
\renewcommand{\arraystretch}{1.08}
\begin{tabularx}{0.92\linewidth}{@{}l>{\centering\arraybackslash}Xcc@{}}
\toprule
\textbf{Model} & \textbf{Harness} & \textbf{M (\%) $\uparrow$} & \cellcolor[HTML]{EAF2FC}\textbf{S (\%) $\uparrow$} \\
\midrule
\multicolumn{4}{@{}l}{\textcolor[HTML]{666666}{\textit{Same-scale baselines}}} \\
Qwen3.6-35B-A3B & Claude Code & 40.0 (4/10) & \cellcolor[HTML]{EAF2FC}0.0 (0/10) \\
Nex-N2.5-mini & Claude Code & 80.0 (8/10) & \cellcolor[HTML]{EAF2FC}20.0 (2/10) \\
Ornith-1.5-35B-A3B & Claude Code & 60.0 (6/10) & \cellcolor[HTML]{EAF2FC}30.0 (3/10) \\
\midrule
\multicolumn{4}{@{}l}{\textcolor[HTML]{666666}{\textit{Larger frontier models}}} \\
DeepSeek-V4-Pro Preview & Claude Code & 60.0 (6/10) & \cellcolor[HTML]{EAF2FC}20.0 (2/10) \\
MiniMax M3 & Claude Code & 70.0 (7/10) & \cellcolor[HTML]{EAF2FC}30.0 (3/10) \\
GPT-5.5 & Codex & \textbf{100.0} (10/10) & \cellcolor[HTML]{EAF2FC}40.0 (4/10) \\
GLM-5.3 & Claude Code & \underline{90.0} (9/10) & \cellcolor[HTML]{EAF2FC}\underline{50.0} (5/10) \\
Kimi-K3 & Claude Code & \underline{90.0} (9/10) & \cellcolor[HTML]{EAF2FC}\textbf{70.0} (7/10) \\
\midrule
\rowcolor{methodhighlight}
\textbf{\sname{} (Ours)} & Claude Code & 70.0 (7/10) & \cellcolor{methodhighlight}\underline{50.0} (5/10) \\
\bottomrule
\end{tabularx}
\endgroup
\end{table}

\noindent\textbf{On NatureBench Lite, \sname{} improves scientific task performance over the base model.}
Table~\ref{tab:naturebench_results} shows that, with Claude Code, \sname{} raises the Surpass-SOTA rate from 0.0\% to 50.0\%, exceeding Nex-N2.5-mini at 20.0\% and matching GLM-5.3 at 50.0\%.
Its Match-SOTA rate also improves from 40.0\% to 70.0\%.
These results show that post-training improves scientific task optimization, enabling \sname{} to match or exceed published reference results on more tasks.

\subsection{Ablation Studies}

\begin{table}[!htbp]
\begin{minipage}[t]{0.43\linewidth}
\vspace{0pt}
\centering
\caption{Effect of post-training stages on MLE-Bench Lite with the \sname{} Harness.}
\label{tab:posttraining_stages}
\begingroup
\footnotesize
\renewcommand{\arraystretch}{1.12}
\setlength{\tabcolsep}{4pt}
\edef\stagerowgap{\the\dimexpr(\ht\strutbox+\dp\strutbox)*28/100\relax}
\begin{tabularx}{\linewidth}{@{}Xr@{}}
\toprule
Training stage & Any Medal (\%) $\uparrow$ \\
\midrule
Base & 37.9 \\[\stagerowgap]
+ General Agent SFT & 47.0 \\[\stagerowgap]
+ MLE SFT & \underline{62.1} \\[\stagerowgap]
\rowcolor{methodhighlight}
\textbf{+ WIRL} & \textbf{72.7} \\[\stagerowgap]
\bottomrule
\end{tabularx}
\endgroup

\end{minipage}\hfill
\begin{minipage}[t]{0.54\linewidth}
\vspace{0pt}
\centering
\caption{RL comparisons on MLE-Bench Lite. Methods are initialized from the same MLE-SFT model.}
\label{tab:rl_comparisons}
\begingroup
\footnotesize
\renewcommand{\arraystretch}{1.12}
\setlength{\tabcolsep}{4pt}
\begin{tabularx}{\linewidth}{@{}Xr@{}}
\toprule
Method & Any Medal (\%) $\uparrow$ \\
\midrule
GRPO & 56.1 \\
PPO-style & 60.6 \\
WIRL w/o trend reward & \underline{66.7} \\
WIRL w/o policy correction & 65.2 \\
\rowcolor{methodhighlight}
\textbf{WIRL} & \textbf{72.7} \\
\bottomrule
\end{tabularx}
\endgroup

\end{minipage}
\end{table}
\noindent\textbf{Effect of post-training stages.}
Table~\ref{tab:posttraining_stages} compares successive training stages under the \sname{} Harness. General Agent SFT raises the average medal rate from 37.9\% to 47.0\%, MLE-domain SFT further increases it to 62.1\%, and WIRL reaches 72.7\%.
Each stage provides an additional gain. In particular, hierarchical SFT substantially improves MLE performance over the base model, while WIRL further improves performance beyond teacher-supervised training through online execution feedback.

\noindent\textbf{RL baselines and key WIRL design choices.}
We compare WIRL with trajectory-level GRPO and PPO-style optimization under the same MLE environment and MLE-SFT initialization. GRPO uses group-relative advantages, whereas PPO-style optimization uses a clipped objective weighted by terminal trajectory rewards without an additional critic. We also evaluate WIRL variants that remove the trend reward or policy correction separately. As shown in Table~\ref{tab:rl_comparisons}, GRPO and PPO-style optimization achieve 56.1\% and 60.6\%, respectively, both below the 62.1\% achieved by MLE-SFT, supporting WIRL's overall design of learning from execution-window feedback. In contrast, the two WIRL variants reach 66.7\% and 65.2\%, respectively, while full WIRL further improves the medal rate to 72.7\%, indicating that both the local trend reward and policy correction contribute to the final improvement. Further analysis in Appendix~\ref{app:rl_waiting} shows that asynchronous execution-window collection reduces the learner waiting fraction compared with trajectory-based GRPO.

\subsection{Case Study: Long-Horizon Optimization}
\label{sec:long_horizon_analysis}

\begin{figure}[!htbp]
    \centering
    \includegraphics[width=\linewidth]{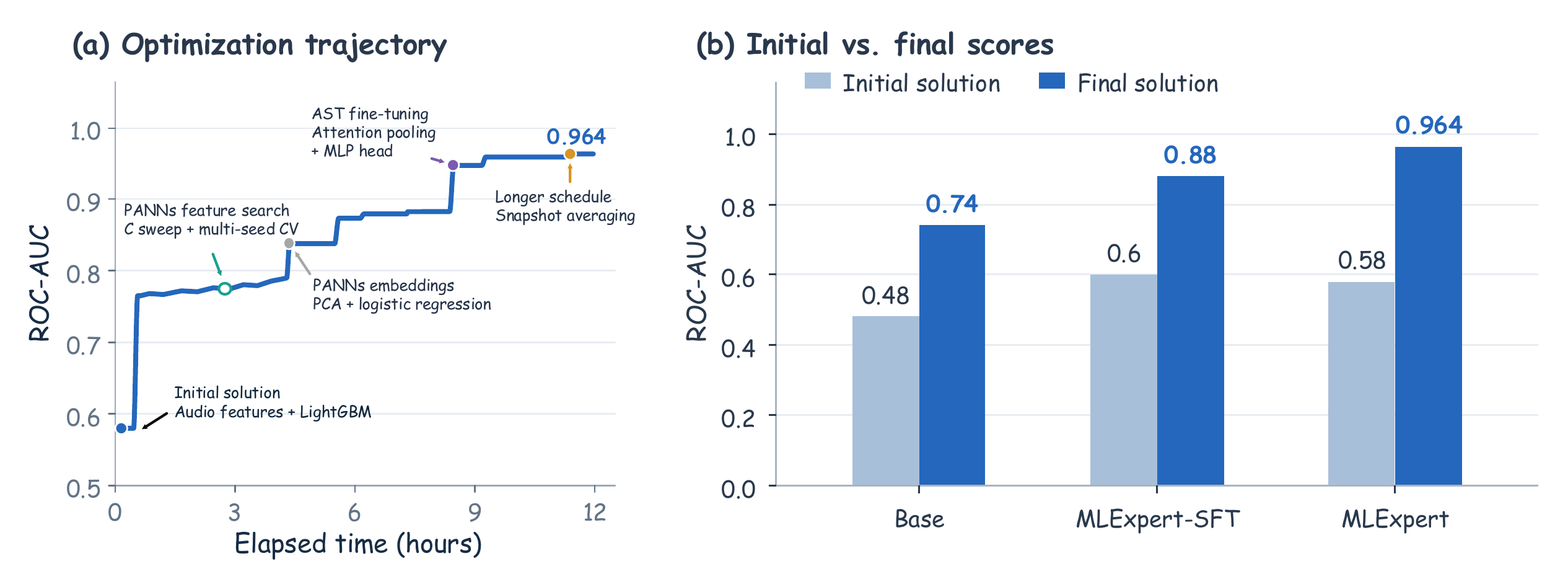}
    \caption{Long-horizon optimization on MLSP 2013 Birds. (a) Representative optimization progress of \sname{}. (b) Initial and final ROC-AUC for Base, \sname{}-SFT, and \sname{}.}
    \label{fig:long_horizon_analysis}
\end{figure}

To examine the long-horizon optimization behavior of \sname{}, we select MLSP 2013 Birds as a representative case. Figure~\ref{fig:long_horizon_analysis}(a) traces the optimization process from audio features and LightGBM through PANNs representations to AST fine-tuning with attention pooling and an MLP head, followed by training-schedule adjustments and snapshot averaging. As shown in Figure~\ref{fig:long_horizon_analysis}(b), \sname{} and \sname{}-SFT start with solutions of similar quality, but \sname{} improves ROC-AUC from 0.58 to 0.964, compared with 0.60 to 0.88 for \sname{}-SFT. This case shows how \sname{} continues to improve solutions through sustained experimentation and execution feedback.

\subsection{Generalization to Other Optimization Tasks}

\noindent\textbf{Agent scaffold optimization.}
Following ShinkaEvolve~\citep{lange2026shinkaevolve}, we optimize a Python scaffold for a fixed GPT-4.1-nano solver on AIME 2025~\citep{zhang2025american} and evaluate it on AIME problems from years not used for optimization.
Table~\ref{tab:aime_transfer} shows that the scaffold optimized by \sname{} achieves 43.7\% held-out accuracy, compared with 23.3\% for the scaffold optimized by the base model.
These results show that post-training improves the ability to design effective agent workflows, with optimization gains extending to unseen problems.
Full results and evaluation protocols are provided in Appendix~\ref{app:aime_details}.

\begin{table}[!htbp]
    \centering
    \caption{AIME scaffold optimization with GPT-4.1-nano. Higher accuracy (\%) and fewer calls are better. Bold/underline: best/second-best.}
    \label{tab:aime_transfer}
    \begingroup
    \small
    \setlength{\tabcolsep}{8pt}
    \renewcommand{\arraystretch}{1.2}
    \begin{tabularx}{0.85\linewidth}{@{}Xrrr@{}}
        \toprule
        Method & Dev. $\uparrow$ & Held-out $\uparrow$ & Calls $\downarrow$ \\
        \midrule
        Initial & 21.1 & 20.0 & \textbf{1.00} \\
        Majority@5 & 24.4 & 24.8 & 5.00 \\
        \midrule
        Base & 28.9 & 23.3 & 5.74 \\
        Kimi-K3 & 45.6 & 40.0 & 8.65 \\
        GLM-5.3 & \textbf{47.8} & \textbf{48.1} & 9.39 \\
        \rowcolor[HTML]{FFFAF0}
        \sname{}-SFT (Ours) & 37.8 & 38.1 & \underline{3.51} \\
        \rowcolor{methodhighlight}
        \sname{} (Ours) & \underline{46.7} & \underline{43.7} & 8.40 \\
        \bottomrule
    \end{tabularx}
    \endgroup

\end{table}

\begin{figure}[!htbp]
    \centering
    \includegraphics[width=\linewidth]{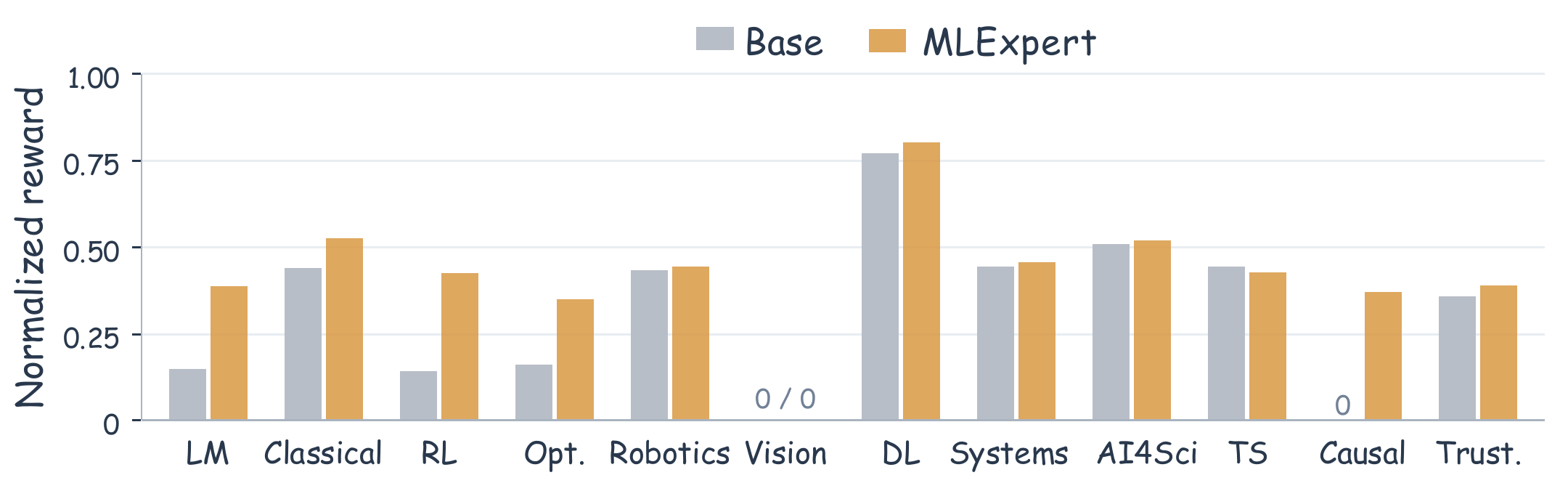}
    \caption{Algorithm-component optimization on 12 selected MLS-Bench-Lite tasks. Bars show normalized verifier rewards for Base and \sname{}.}
    \label{fig:mls_transfer}
\end{figure}

\noindent\textbf{Algorithm-component optimization on MLS-Bench-Lite.}
On MLS-Bench-Lite~\citep{lyu2026mls}, we optimize algorithmic components within fixed research pipelines on 12 selected tasks spanning 12 domains, under the same five-hour budget.
Figure~\ref{fig:mls_transfer} shows that \sname{} improves the mean verifier reward from 0.322 to 0.425, outperforming Base on ten tasks, tying on one, and scoring lower on one.
The gains span language-model demasking, multi-objective optimization, and causal discovery, demonstrating transfer beyond end-to-end MLE to component optimization across scientific domains.
Full protocols and task-level results are provided in Appendix~\ref{app:mls_details}.
\FloatBarrier

\section{Related work}

\subsection{Agents for Automated ML Research}
\label{sec:rw_automated_ml}
LLM agents increasingly automate data analysis, model development, and experimental iteration, with recent work extending to autonomous AI system design and improvement~\citep{wu2026ai4ai}.
AIDE~\citep{aide} and MLE-STAR~\citep{mle-star} improve candidate solutions through iterative search and refinement, while MLEvolve~\citep{du2026mlevolve} and MARS~\citep{chen2026mars} use structured tree- or graph-based search, memory, and planning to organize long-horizon experimentation.
R\&D-Agent~\citep{rdagent} studies multi-role collaboration, while InternAgent~\citep{feng2026internagent} and FM-Agent~\citep{li2025fm} extend autonomous AI research across machine learning, algorithm optimization, and engineering tasks.
Recent systems including Arbor~\citep{jin2026arbor} and Argus~\citep{li2026argus} maintain persistent research states and increase model control.

Recent work develops MLE capabilities by training models on interaction experience.
ML-Agent~\citep{liu2025mlagent} combines multi-turn demonstrations with single-step RL, while RL for MLE Agents~\citep{yang2026rl4agents}, MLE-RL~\citep{li2026mlerl}, and AceGRPO~\citep{cai2026acegrpo} study online MLE training with execution feedback, asynchronous sampling, or intermediate trajectory states.
Learning to Ideate~\citep{zhang2026learning} trains models to generate algorithm improvement suggestions, whereas Frontis-MA1~\citep{yang2026frontis} applies SFT and RL to program evolution operators and combines them through an external evolutionary framework.
These approaches primarily train individual interaction steps or local components of externally organized optimization workflows.
SandMLE~\citep{zhou2026sandmle} applies trajectory-wise on-policy RL in micro-scale synthetic MLE environments.
We train model-directed long-horizon interaction policies on real-world MLE tasks, with WIRL using execution windows to learn multi-turn decisions from delayed execution feedback under asynchronous updates.

\subsection{Post-Training Long-Horizon Agentic Models}
\label{sec:rw_long_horizon_agents}

Recent research has increasingly focused on post-training agentic models for sustained interaction, complex tool use, and long-horizon decision making.
SWE-Gym~\citep{pan2024swe-gym} trains coding agents using trajectories collected from executable environments, while Search-R1~\citep{jin2025searchr1} and WebRL~\citep{qi2025webrl} learn multi-turn tool use from environment feedback and outcome rewards.
AgentFlow~\citep{li2026flow} directly optimizes planning policies in general multi-turn agentic interaction and uses trajectory-level outcomes to train intermediate decisions. Recent work has also explored generalist agentic models for long-horizon interaction.
Agents-A1~\citep{bai2026agentsa1} uses long-horizon knowledge-action trajectories and multi-stage training, together with multi-teacher on-policy distillation (OPD), to combine capabilities from multiple domain teachers into a unified model.
Ornith-1.5~\citep{ornith_1_5} extends self-scaffolding by jointly optimizing task generation, scaffold construction, and solution rollouts through RL.

\section{Conclusion}

In this work, we introduce \sname{}, a 35B MoE agentic model trained to autonomously conduct and improve machine learning experiments over long interaction horizons.
We build an automated MLE task factory and a model-directed interaction harness, and train \sname{} with hierarchical agentic SFT and WIRL to learn from teacher demonstrations and execution feedback collected from its own exploration.
Extensive experiments show that \sname{} improves over its base model on MLE-Bench Lite and achieves performance comparable to larger frontier models.
Further improvements on NatureBench Lite and in agent scaffold and algorithm optimization demonstrate cross-task generalization.
These results support treating long-horizon optimization decision making as a trainable model capability, providing a practical path toward autonomous machine learning engineering and scientific experimentation.

\clearpage

\bibliographystyle{m1_numeric}
\bibliography{references}

\clearpage
\appendix

\section{Details of the Kaggle-MLE Factory}
\label{app:kaggle_factory}

We begin with a Meta-Kaggle\footnote{\url{https://www.kaggle.com/datasets/kaggle/meta-kaggle}} competition metadata snapshot containing 11,649 entries.
Guided by the task-selection principles of MLE-bench~\citep{mle-bench}, the discovery agent considers task relevance, data availability, and the feasibility of offline evaluation.
Automated initial screening uses competition completion, at least 50 participating teams, and a dataset size limit of 200\,GB as default criteria, checks submission-interface compatibility, and consults an exclusion list of existing tasks.
The agent further selects candidates across domains and data modalities, including vision, text, audio, tabular data, and time series.
Candidates are collected and processed in multiple batches.

The downloader and scraper retrieve competition data, official descriptions, and leaderboard information.
The analyzer agent extracts data structures, field information, and modality characteristics to form a task specification, which the generator agent uses to produce data-preparation and scoring code.
When test labels are unavailable, the pipeline constructs a held-out split from the original labeled data and keeps its labels on the evaluation side.
The executor runs generated task packages on real data, while the validator agent checks structural completeness, submission-format consistency, non-degenerate labels, train--test identifier overlap, and data availability.
The validator also executes the scoring function to check that it produces a valid score.
Execution and validation errors are returned to the construction agents for repair.
We conduct a final human review of the 160 registered tasks, excluding six with unresolved label or scoring-implementation issues and retaining 154 tasks for subsequent trajectory collection.

\section{Interaction Harness Tools}
\label{app:harness_tools}

The Interaction Harness described in Section~\ref{sec:harness} provides tools for inspecting the task environment, revising candidate solutions, running experiments, retrieving previous states, and managing submissions.
The model selects subsequent actions based on execution feedback; Table~\ref{tab:harness_tools} summarizes the available tools and their purposes.

\begin{table}[htbp]
\centering
\caption{Tools provided by the Interaction Harness.}
\label{tab:harness_tools}
\small
\setlength{\tabcolsep}{5pt}
\begin{tabular}{@{}>{\raggedright\arraybackslash}p{0.52\linewidth} p{0.44\linewidth}@{}}
\toprule
Tools & Purpose \\
\midrule
\texttt{analyze} &
Delegate data, result, or error analysis \\

\texttt{execute\_bash} &
Inspect data and runtime environment \\

\texttt{write\_full\_code}, \texttt{patch\_code} &
Create or revise candidate solutions \\

\texttt{execute\_code}, \texttt{control\_run} &
Launch and control experiments \\

\texttt{write\_plan}, \texttt{edit\_plan}, \texttt{read\_plan} &
Maintain persistent experiment plans \\

\texttt{list\_nodes}, \texttt{select\_node} &
Retrieve candidate states and history \\

\texttt{invalidate\_node} &
Exclude candidates with untrustworthy metrics \\

\texttt{update\_answer}, \texttt{get\_current\_answer} &
Maintain valid submission candidates \\
\bottomrule
\end{tabular}
\end{table}

\section{Agentic MLE Post-Training Details}
\label{app:posttraining_details}

\subsection{Details of Hierarchical Agentic SFT}
\label{app:sft_details}

\noindent Training data sources.
The first SFT stage uses general coding and tool-use trajectories from StepFun~\citep{huang2026step}, while the second stage combines MLE teacher trajectories with literature-derived scientific trajectories. The MLE tasks comprise 154 tasks constructed by the Kaggle-MLE Factory and 134 tasks selected from MLE-Dojo~\citep{qiang2026mledojo}, with competitions overlapping with MLE-Bench removed. Following the task-construction approach of NatureBench~\citep{wang2026naturebench}, we construct scientific experimentation tasks from 164 papers published in Nature-family journals, covering biology, materials science, physics, chemistry, and medicine, among other fields. The data used in these scientific tasks do not overlap with the NatureBench evaluation set. MLE teacher trajectories are generated by GLM-5.2~\citep{glm_5_2}, while scientific experimentation trajectories are generated by GLM-5.2 and Qwen3.8-Max~\citep{qwen_3_8_max}.

\noindent Retained corpus and sampling.
The corpus supplied to MLE-domain SFT contains 636 retained teacher trajectories: 412 MLE trajectories covering 236 tasks and 224 scientific experimentation trajectories covering 77 tasks.
Of the scientific trajectories, 143 are generated by GLM-5.2 and 81 by Qwen3.8-Max.
We count trajectories by distinct source runs before segmentation and repeated sampling; main-agent segments, initial-solution prefixes, and analysis sub-agent interactions from the same run share a single source-trajectory count.
The task counts describe coverage of the retained corpus within the task pools above.
Table~\ref{tab:sft_corpus_statistics} summarizes the source trajectories and training input records.

\begin{table}[htbp]
\centering
\caption{Composition of the MLE-domain SFT input corpus. Source trajectories count distinct retained runs. Input records count serialized training examples after segmentation and repeated sampling, before training-time preprocessing and packing.}
\label{tab:sft_corpus_statistics}
\small
\setlength{\tabcolsep}{8pt}
\begin{tabular}{@{}llrr@{}}
\toprule
Domain & Teacher & Source trajectories & Input records \\
\midrule
MLE & GLM-5.2 & 412 & 4,961 \\
Scientific experimentation & GLM-5.2 & 143 & 342 \\
Scientific experimentation & Qwen3.8-Max & 81 & 236 \\
\midrule
Total & & 636 & 5,539 \\
\bottomrule
\end{tabular}
\end{table}

The MLE input files contain 1,559 main-agent records and 1,843 analysis sub-agent records.
The main-agent file is included twice and the sub-agent file once, contributing 4,961 records to the input mixture.
The scientific data contain 171 GLM-5.2 records and 118 Qwen3.8-Max records before repetition; each is included twice, contributing 342 and 236 records, respectively.
The Qwen3.8-Max data combine two collection batches containing 49 and 32 source trajectories, which yield 66 and 52 records before repetition.
These sampling multiplicities determine the input mixture, while the trajectory counts in Table~\ref{tab:sft_corpus_statistics} count each source run once.

After training-time preprocessing and sequence packing, the training log records 1,173 sequences with a mean length of approximately 257,843 tokens, corresponding to approximately 302.45 million tokens in the packed dataset.
This token count includes conditioning context and generated outputs and describes the dataset before repetition across the eight training epochs.
The loss is computed on the output positions specified below.

\noindent\textbf{Outcome-guided supervision selection.}
For MLE trajectories with a valid leaderboard reference, let $(b_1,\ldots,b_K)$ denote the valid scores obtained in submission order. Here, $b_j\in[0,1]$ is the submission's relative standing on the task's private leaderboard, accounting for the direction of the evaluation metric so that larger values indicate better performance.
We use the same quality threshold $\eta=0.4$ for trajectory filtering and initial-solution extraction.
Trajectories with $\max_j b_j<\eta$ are discarded entirely; those with $\max_j b_j\geq\eta$ pass the quality filter.
For retained trajectories, we define the first occurrence of the best score as
\[
j^\star=\min\arg\max_j b_j,
\]
and measure the average relative degradation after this point as
\begin{equation}
\label{eq:sft_degradation}
d(\tau)=
\begin{cases}
\displaystyle
\frac{
b_{j^\star}
-
\frac{1}{K-j^\star}
\sum_{j=j^\star+1}^{K} b_j
}{
b_{j^\star}
},
& j^\star<K,\\[6pt]
0,
& j^\star=K.
\end{cases}
\end{equation}

Given a degradation threshold $\delta\geq0$, we use the first best-performing submission as the truncation anchor when $d(\tau)>\delta$; otherwise, we use the last submission.
The retained prefix includes the anchor's \texttt{update\_answer} call and its tool response, followed by a randomly selected extension of up to ten complete tool interactions.
The extension stops before the next submission and ends after a tool response.
This filtering removes degraded trajectory tails according to their overall post-peak behavior.

The default degradation threshold is $\delta=0.05$, as reported in Table~\ref{tab:training_hyperparameters}.

\noindent\textbf{Supervision Samples and Training Objective.}
Let $(w_1,\ldots,w_S)$ denote the supervision samples constructed from a trajectory, and let $x_{s,t}$ be the $t$-th token in sample $w_s$.
The SFT objective is
\begin{equation}
\label{eq:sft_objective}
\mathcal{L}_{\mathrm{SFT}}(\theta)
=
-\sum_{s=1}^{S}
\sum_{t\in\mathcal{M}(w_s)}
\log \pi_\theta
\left(
x_{s,t}\mid x_{s,<t}
\right),
\end{equation}
where $\mathcal{M}(w_s)$ contains only the positions of newly generated reasoning, text responses, and tool-call tokens in the current segment.
Tool feedback and inherited interaction history are used only as conditioning context and are masked out from the loss.

\subsection{Details of Windowed Interaction Reinforcement Learning}
\label{app:wirl_details}

\noindent\textbf{Local Optimization Objective.}
Let $H_m$ denote the experimental history at the start of execution window $u_m$ and $\zeta_m$ the closed-loop interaction segment within the window.
The conditional distribution $P_\theta(\cdot\mid H_m)$ is induced by the policy and the interaction environment, and $R_m(\zeta_m)$ is the reward computed from the corresponding execution result.
For a fixed distribution $D$ over window-start histories, we define the local objective as
\begin{equation}
\label{eq:window_objective}
J_{\mathrm{win}}(\theta;D)
=
\mathbb{E}_{H_m\sim D}
\mathbb{E}_{\zeta_m\sim P_\theta(\cdot\mid H_m)}
\left[R_m(\zeta_m)\right].
\end{equation}

\noindent\textbf{Local Performance Trend.}
For the first valid submission, we set $T_m=0$.
For subsequent valid submissions, let $s_m$ and $s_{\mathrm{prev}}$ denote the current score and the most recent valid score, respectively.
We define the score change according to the metric direction as
\begin{equation}
\label{eq:local_score_trend}
\Delta_m=
\begin{cases}
s_m-s_{\mathrm{prev}}, & \text{higher is better},\\
s_{\mathrm{prev}}-s_m, & \text{lower is better},
\end{cases}
\qquad
T_m=
\begin{cases}
0.2, & \Delta_m>0,\\
0, & \Delta_m=0,\\
-0.2, & \Delta_m<0.
\end{cases}
\end{equation}

\noindent\textbf{Rollout Log-Probability and Policy Correction Details.}
For each generated output token, we record the corresponding rollout log-probability $\ell^{\mathrm{roll}}_{mjt}$ during data collection.
The correction ratio $q_{mjt}$ in Eq.~\ref{eq:policy_correction} compares this recorded value with the pre-update training log-probability $\ell^{\mathrm{old}}_{mjt}$.
The bounds $q_{\min}$ and $q_{\max}$ define the range in which token-level policy correction is applied.
Accordingly, the correction is used at the token level rather than as a trajectory-level importance-sampling ratio.




\noindent\textbf{Token-Level Policy-Gradient Loss.}
For output token $t$ from model call $j$ in window $u_m$, we define the probability ratio $r_{mjt}=\exp(\ell_{\theta,mjt}-\ell^{\mathrm{old}}_{mjt})$ between the current and pre-update policies.
Using the shared window weight $\widehat A_m=R_m$, the PPO-style token-level policy-gradient loss is
\begin{equation}
\label{eq:policy_loss}
L^{\mathrm{PG}}_{mjt}
=
\max\!\left(
-r_{mjt}\widehat A_m,\;
-\operatorname{clip}(r_{mjt},1-\epsilon_-,1+\epsilon_+)\widehat A_m
\right),
\end{equation}
where $\epsilon_-$ and $\epsilon_+$ control the lower and upper clipping bounds, respectively.
This loss is weighted by $c_{mjt}$ in the overall objective in Eq.~\ref{eq:wirl_objective}.

\noindent\textbf{Window Normalization and Masking Details.}
The correction weight $c_{mjt}$ is applied only to the policy-gradient term and does not modify the KL regularization, entropy term, or window-normalization denominator.
The training mask $M_{mjt}$ selects newly generated main-model output tokens.
For a window with no valid output tokens, the masked token sum is zero, while $\max(N_m,1)$ prevents division by zero.
Padded window slots therefore contribute zero loss, and the outer normalization remains over the $B$ window slots in the batch.


\noindent\textbf{Execution Feedback Filtering.}
WIRL updates use only windows with trusted execution feedback.
Failures caused by infrastructure errors or unavailable scoring signals are excluded from training rather than being assigned the model-error reward in Eq.~\ref{eq:window_reward}.


\subsection{Hyperparameters}
\label{app:hyperparameters}

Table~\ref{tab:training_hyperparameters} summarizes the key SFT training settings and the WIRL \texttt{medal\_absolute\_v1} configuration corresponding to the reward formulation in Section~\ref{sec:wirl}.
The supervision thresholds are the defaults of the trajectory-filtering pipeline.
Both SFT stages and WIRL use cosine learning-rate decay; WIRL uses no learning-rate warmup.

\begin{table}[htbp]
    \centering
    \caption{Key hyperparameters for hierarchical agentic SFT and WIRL.}
    \label{tab:training_hyperparameters}
    \begingroup
    \small
    \setlength{\tabcolsep}{4pt}
    \renewcommand{\arraystretch}{1.12}
    \begin{tabularx}{\linewidth}{@{}lLr@{}}
        \toprule
        \textbf{Hyperparameter} & \textbf{Description} & \textbf{Value} \\
        \midrule
        \multicolumn{3}{@{}l}{\textit{General Agent SFT}} \\
        \arrayrulecolor{black!30}\cmidrule(lr){1-2}\arrayrulecolor{black}
        Learning rate & Initial / minimum learning rate & $2\times10^{-5}\,/\,10^{-6}$ \\
        Batch size & Global batch size & 32 \\
        Sequence length & Maximum training tokens per sequence & 131,072 \\
        Training epochs & Number of passes over the dataset & 1 \\
        Warmup ratio & Fraction of training steps for warmup & 0.05 \\
        \midrule
        \multicolumn{3}{@{}l}{\textit{MLE SFT}} \\
        \arrayrulecolor{black!30}\cmidrule(lr){1-2}\arrayrulecolor{black}
        Learning rate & Initial / minimum learning rate & $10^{-5}\,/\,10^{-6}$ \\
        Batch size & Global batch size & 8 \\
        Sequence length & Maximum training tokens per sequence & 262,144 \\
        Training epochs & Number of passes over the dataset & 8 \\
        Warmup ratio & Fraction of training steps for warmup & 0.05 \\
        $\eta$ & Supervision quality threshold & 0.4 \\
        $\delta$ & Post-peak degradation threshold & 0.05 \\
        \midrule
        \multicolumn{3}{@{}l}{\textit{WIRL}} \\
        \arrayrulecolor{black!30}\cmidrule(lr){1-2}\arrayrulecolor{black}
        Learning rate & Initial / minimum learning rate & $10^{-6}\,/\,10^{-7}$ \\
        $B$ & Execution windows per training batch & 32 \\
        Sequence length & Maximum training tokens per sequence & 131,072 \\
        Sampling & Rollout temperature / top-$p$ / top-$k$ & $0.6\,/\,0.95\,/\,20$ \\
        $[q_{\min},q_{\max}]$ & Token-level policy correction interval & $[0.2,\,4.0]$ \\
        $(\epsilon_-,\epsilon_+)$ & Lower / upper policy clipping parameters & $(0.3,\,3.0)$ \\
        $\lambda_{\mathrm{KL}}$ & KL regularization coefficient & $10^{-3}$ \\
        $\lambda_H$ & Entropy regularization coefficient & $10^{-4}$ \\
        \bottomrule
    \end{tabularx}
    \endgroup
\end{table}

\section{Supplementary Experimental Protocols and Results}
\label{app:experiment_details}

\subsection{MLE-Bench Lite Evaluation Details}
\label{app:mle_results}

Table~\ref{tab:mle_seed_results} reports MLE-Bench Lite results with the \sname{} Harness, including average rates and SEM over three evaluations for \sname{} and same-scale baselines, and one result for each larger frontier model.
All rates are computed over 22 tasks; tasks without a valid scored submission count as no medal.
For the general-purpose harness results in Table~\ref{tab:mle_main_results}, \sname{} and the same-scale baselines are each evaluated three times with Claude Code. Any Medal rates (mean $\pm$ SEM) are $42.42\% \pm 4.01\%$ for Qwen3.6-35B-A3B, $46.97\% \pm 3.03\%$ for Ornith-1.5-35B-A3B, $36.36\% \pm 6.94\%$ for Nex-N2.5-mini, and $69.70\% \pm 1.52\%$ for \sname{}. Each larger frontier model has one reported evaluation result.
The Kimi-K3 result with Claude Code\footnote{\url{https://code.claude.com/docs/en/overview}} and the GPT-5.6 Sol result with Codex\footnote{\url{https://learn.chatgpt.com/docs/codex/cli}} are from Frontis-MA1~\citep{yang2026frontis}; the remaining results are from our evaluations.

\begin{table}[htbp]
\centering
\caption{MLE-Bench Lite results with the \sname{} Harness. Percentages are reported as average $\pm$ SEM over three evaluations for \sname{} and same-scale baselines; one result is reported for each larger frontier model.}
\label{tab:mle_seed_results}
\begingroup
\small
\setlength{\tabcolsep}{3pt}
\renewcommand{\arraystretch}{1.15}
\begin{tabular*}{\linewidth}{@{\extracolsep{\fill}}lccccc@{}}
\toprule
\textbf{Model} & \textbf{Med+ (\%)} & \textbf{Gold (\%)} & \textbf{Silver (\%)} & \textbf{Bronze (\%)} & \textbf{Any Medal (\%)} \\
\midrule
Qwen3.6-35B-A3B & $62.12 \pm 4.01$ & $22.73 \pm 2.62$ & $7.58 \pm 4.01$ & $\underline{7.58} \pm 1.52$ & $37.88 \pm 5.46$ \\
Ornith-1.5-35B-A3B & $57.58 \pm 1.52$ & $22.73 \pm 6.94$ & $12.12 \pm 1.52$ & $4.55 \pm 2.62$ & $39.39 \pm 4.01$ \\
Nex-N2.5-mini & $66.67 \pm 1.52$ & $24.24 \pm 6.06$ & $13.64 \pm 2.62$ & $\mathbf{9.09} \pm 4.55$ & $46.97 \pm 1.52$ \\
\midrule
\makecell[l]{DeepSeek-V4-Pro\\Preview} & 86.36 & 40.91 & 13.64 & \textbf{9.09} & 63.64 \\
MiniMax M3 & 81.82 & 54.55 & 13.64 & 0.00 & 68.18 \\
GLM-5.3 & \underline{90.91} & \underline{59.09} & 13.64 & 0.00 & \underline{72.73} \\
Kimi-K3 & \textbf{95.45} & \textbf{63.64} & 9.09 & 0.00 & \underline{72.73} \\
GPT-5.5 xhigh & 81.82 & 45.45 & \textbf{22.73} & 4.55 & \underline{72.73} \\
GPT-5.6 Sol & \underline{90.91} & \underline{59.09} & 13.64 & 4.55 & \textbf{77.27} \\
\midrule
\sname{} & $84.85 \pm 3.03$ & $50.00 \pm 2.62$ & $\underline{16.67} \pm 1.52$ & $6.06 \pm 1.52$ & $\underline{72.73} \pm 2.62$ \\
\bottomrule
\end{tabular*}
\endgroup
\end{table}

\noindent\textbf{Medal composition and harness compatibility.}
Table~\ref{tab:mle_seed_results} shows that post-training improves both medal coverage and the frequency of higher-tier outcomes: the Gold rate rises from 22.73\% for the base model to 50.00\% for \sname{}, while Any Medal rises from 37.88\% to 72.73\%.
Under Claude Code, \sname{} retains a substantial advantage over the base model, with 69.7\% versus 42.4\% Any Medal.
Together, these results support improvements in solution quality that remain effective across both task-specific and general-purpose interaction harnesses.

The post-training stage results in Table~\ref{tab:posttraining_stages} are averaged over three evaluations. Any Medal rates (mean $\pm$ SEM) are $37.88\% \pm 5.46\%$ for Base, $46.97\% \pm 1.52\%$ for General Agent SFT, $62.12\% \pm 1.52\%$ for MLE SFT, and $72.73\% \pm 2.62\%$ for WIRL.
Table~\ref{tab:wirl_ablation} expands the RL comparisons in Table~\ref{tab:rl_comparisons}, reporting average $\pm$ SEM over three evaluations.

\begin{table}[htbp]
\centering
\caption{RL baselines and WIRL ablations on MLE-Bench Lite. Any Medal rates are reported as average $\pm$ SEM over three evaluations.}
\label{tab:wirl_ablation}
\begingroup
\small
\setlength{\tabcolsep}{12pt}
\renewcommand{\arraystretch}{1.12}
\begin{tabular}{@{}lc@{}}
\toprule
\textbf{Method} & \textbf{Any Medal (\%)} \\
\midrule
GRPO & $56.06 \pm 5.46$ \\
PPO-style & $60.61 \pm 4.01$ \\
WIRL w/o trend reward & $\underline{66.67} \pm 3.03$ \\
WIRL w/o policy correction & $65.15 \pm 5.46$ \\
\rowcolor{methodhighlight}
\textbf{WIRL} & $\mathbf{72.73} \pm 2.62$ \\
\bottomrule
\end{tabular}
\endgroup
\end{table}

\FloatBarrier

\begin{figure}[!t]
    \centering
    \includegraphics[width=0.65\linewidth]{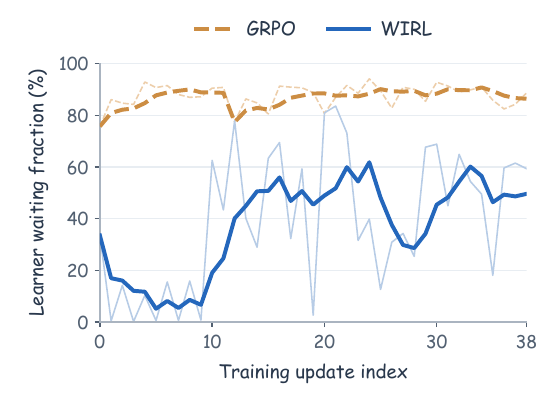}
    \caption{Learner waiting fractions during RL training. Light curves show recorded per-update values; dark curves show trailing five-update means.}
    \label{fig:rl_learner_waiting}
\end{figure}

\phantomsection
\label{app:rl_waiting}
\noindent \textbf{Learner waiting time during RL training.}
Figure~\ref{fig:rl_learner_waiting} compares the learner waiting fractions of GRPO and WIRL under the same training setup. The waiting fraction is the time spent waiting for training data divided by the sum of waiting and training time. Over the recorded updates, the mean per-update waiting fraction is 87.6\% for GRPO and 39.3\% for WIRL, a difference of 48.3 percentage points. WIRL exhibits substantially lower learner waiting overhead during RL training.

\subsection{NatureBench Lite}
\label{app:naturebench_details}

We use the ten-task NatureBench Lite subset listed by Frontis-MA1~\citep{yang2026frontis}, with the S and M definitions from NatureBench~\citep{wang2026naturebench}.
We follow the official NatureBench evaluation protocol, with a four-hour runtime budget per task that excludes time spent scoring submissions.
Following NatureBench, we compute the normalized relative gap for each evaluation instance by subtracting the paper-reported SOTA value from the agent's primary metric value, adjusting the sign so that improvements are positive, and dividing by the absolute SOTA value.
Instances without a valid submission receive a gap of $-1$.
The task-level score $g$ is the mean gap across instances.
Match-SOTA and Surpass-SOTA report the percentages of tasks with $g\geq0$ and $g>0.1$, respectively.
We report a single evaluation of \sname{} over the ten tasks, with Surpass-SOTA on five tasks and Match-SOTA on seven tasks, corresponding to 50.0\% and 70.0\%, respectively.
DeepSeek-V4-Pro Preview and MiniMax M3~\citep{minimax_m3} results are taken from Table~2 of Frontis-MA1.
Our NatureBench evaluations use NVIDIA H200 GPUs; external reference results retain their original evaluation conditions.

\noindent\textbf{Matching and surpassing scientific reference results.}
Under Claude Code, \sname{} improves Surpass-SOTA from 0.0\% to 50.0\% and Match-SOTA from 40.0\% to 70.0\% relative to the base model.
Compared with Nex-N2.5-mini, \sname{} has a higher Surpass-SOTA rate, 50.0\% versus 20.0\%, but a lower Match-SOTA rate, 70.0\% versus 80.0\%.
\sname{} more frequently exceeds the reference by the Surpass-SOTA margin, while Nex matches the reference on a larger fraction of evaluations.
\sname{} matches the 50.0\% Surpass-SOTA rate of GLM-5.3, whose Match-SOTA rate is higher at 90.0\%.


\subsection{AIME Agent Scaffold Optimization}
\label{app:aime_details}

The outer model optimizes a Python agent scaffold that calls a fixed \texttt{gpt-4.1-nano-2025-04-14} solver.
All optimizers start from the same single-query scaffold and share the task environment, tools, and evaluator.
Each search has a four-hour time limit and a maximum of 1,000 interaction steps, with a context limit of 200,000 tokens and an output limit of 65,536 tokens per step.
The scaffold can modify prompting, calling, verification, voting, and answer-extraction procedures, with at most ten inner-model calls per problem.
Initial uses one solver call, while Majority@5 takes a majority vote over five independently sampled answers.

Development uses all 30 problems from AIME 2025, with three evaluations per problem for each candidate scaffold.
The outer model receives aggregate accuracy, call counts, and sanitized error summaries.
Exact-request response caching is used during development to reuse inner-solver responses to identical requests.
After search, the selected scaffold is frozen and evaluated using fresh solver calls on AIME 2023, 2024, and 2026.
Each held-out year contains 30 problems and is evaluated three times; these years provide no feedback for scaffold development or selection.
Thus, each optimizer contributes one frozen scaffold, with three evaluation repetitions per year.

Accuracy is the fraction of answers that exactly match the reference integer answer; missing or incorrectly formatted answers count as incorrect.
Table~\ref{tab:aime_full_results} reports development accuracy, accuracy for each held-out year, the mean across the three held-out years, and mean inner-model calls per problem on the held-out evaluations.
Each annual accuracy averages the three evaluations of that year's 30 problems.

\begin{table}[htbp]
    \centering
    \caption{Complete AIME scaffold optimization results. Accuracy (\%) is higher-is-better; calls are lower-is-better. Held-out accuracy and calls are averaged over AIME 2023, 2024, and 2026; each year's accuracy averages three evaluations of 30 problems. Bold and underline mark the best and second-best distinct values in each column, including ties.}
    \label{tab:aime_full_results}
    \begingroup
    \small
    \setlength{\tabcolsep}{7pt}
    \renewcommand{\arraystretch}{1.15}
    \begin{tabular*}{\linewidth}{@{\extracolsep{\fill}}lrrrrrr@{}}
        \toprule
        \textbf{Method} & \multicolumn{1}{c}{\textbf{Development}} & \multicolumn{4}{c}{\textbf{Held-out accuracy (\%)}} & \makecell[r]{\textbf{Calls per}\\\textbf{problem}} \\
        \cmidrule(lr){2-2}\cmidrule(lr){3-6}
        & 2025 & 2023 & 2024 & 2026 & Average & \\
        \midrule
        Initial & 21.1 & 20.0 & 23.3 & 16.7 & 20.0 & \textbf{1.00} \\
        Majority@5 & 24.4 & 18.9 & 32.2 & 23.3 & 24.8 & 5.00 \\
        \midrule
        Base & 28.9 & 18.9 & 28.9 & 22.2 & 23.3 & 5.74 \\
        Ornith-1.5-35B-A3B & 24.4 & 20.0 & 18.9 & 16.7 & 18.5 & \textbf{1.00} \\
        Nex-N2.5-mini & 32.2 & 17.8 & 27.8 & 33.3 & 26.3 & 5.93 \\
        Kimi-K3 & 45.6 & 37.8 & 41.1 & \underline{41.1} & 40.0 & 8.65 \\
        GLM-5.3 & \textbf{47.8} & \textbf{48.9} & \textbf{51.1} & \textbf{44.4} & \textbf{48.1} & 9.39 \\
        \midrule
        \sname{}-SFT (Ours) & 37.8 & 35.6 & 42.2 & 36.7 & 38.1 & \underline{3.51} \\
        \sname{} (Ours) & \underline{46.7} & \underline{43.3} & \underline{43.3} & \textbf{44.4} & \underline{43.7} & 8.40 \\
        \bottomrule
    \end{tabular*}
    \endgroup
\end{table}

\noindent\textbf{Solver-call efficiency after SFT.}
The scaffold optimized by \sname{}-SFT achieves 38.1\% held-out accuracy with 3.51 solver calls per problem, approaching the 40.0\% achieved by Kimi-K3 with approximately 59\% fewer calls.
It also outperforms Majority@5, which reaches 24.8\% with five calls per problem.
\sname{}-SFT therefore improves held-out accuracy over fixed majority voting while using fewer solver calls, suggesting that its optimized scaffold uses the solver more effectively.
These results indicate that SFT already supports effective scaffold optimization across held-out years.

\noindent\textbf{Further gains from WIRL.}
WIRL raises held-out accuracy from 38.1\% to 43.7\%, a gain of 5.6 percentage points, with improvements on all three held-out years.
The resulting scaffold uses 8.40 calls per problem, compared with 3.51 for \sname{}-SFT and 8.65 for Kimi-K3.
With a similar number of calls to Kimi-K3, \sname{} exceeds its 40.0\% accuracy, while GLM-5.3 achieves 48.1\% with 9.39 calls.
The SFT and RL results therefore offer different accuracy--call trade-offs: SFT yields an effective scaffold with fewer calls, while further RL finds a more accurate scaffold that uses more calls.

\subsection{MLS-Bench-Lite}
\label{app:mls_details}

We evaluate 12 selected tasks spanning 12 domains in MLS-Bench-Lite.
Table~\ref{tab:mls_full_results} lists the complete task set and the normalized rewards for Base and \sname{}.
Both models use the same Agentic MLE harness, task descriptions, interaction tools, and official Harbor task environments.
Each task has a five-hour search budget and a maximum of 2,000 interaction steps, with a context limit of 200,000 tokens and an output limit of 65,536 tokens per step.
CPU, GPU, memory, and verifier time limits follow each official task's resource specification.

The model edits the permitted algorithmic component, runs public experiments, and selects its final candidate through \texttt{update\_answer}.
Final performance is measured by the official hidden Harbor verifier, which aggregates task-specific metrics into a normalized reward.
We report rewards on the zero-to-one scale, with higher values indicating better performance, and compute the mean reward as the arithmetic average over all 12 tasks.
The mean rewards are 0.321543 for Base and 0.425198 for \sname{}.
\sname{} scores higher on ten tasks, ties on one, and scores lower on one.

\begin{table}[htbp]
    \centering
    \caption{Results on 12 tasks selected from MLS-Bench-Lite, one per domain. Scores are normalized rewards from the official Harbor verifier; the final row is the arithmetic mean over the 12 tasks. Bold and underline mark the best and second-best values within each row, including ties.}
    \label{tab:mls_full_results}
    \begingroup
    \small
    \setlength{\tabcolsep}{4pt}
    \renewcommand{\arraystretch}{1.2}
    \begin{tabularx}{\linewidth}{@{}>{\raggedright\arraybackslash}p{0.26\linewidth}Xrr@{}}
        \toprule
        \textbf{Domain} & \textbf{Official task title} & \textbf{Base} & \textbf{\sname{}} \\
        \midrule
        Language Models & Masked Diffusion Demasking Policy & \underline{0.148150} & \textbf{0.386696} \\
        Classical \& Adaptive Learning & Geometry-Robust Clustering Algorithm & \underline{0.439982} & \textbf{0.525322} \\
        Reinforcement Learning & Value-Based Discrete Control & \underline{0.141849} & \textbf{0.425318} \\
        Optimization \& Theory & Multi-Objective Evolutionary Survival and Variation & \underline{0.160624} & \textbf{0.350473} \\
        Robotics & Behavioral Cloning Loss for Manipulation & \underline{0.435532} & \textbf{0.444879} \\
        Vision \& Generation & 3D Scene Densification Strategy & \textbf{0.000000} & \textbf{0.000000} \\
        Deep Learning & Convolutional Activation Nonlinearity & \underline{0.773039} & \textbf{0.804121} \\
        ML Systems \& Efficient ML & Post-Training Weight Quantization & \underline{0.443933} & \textbf{0.456671} \\
        AI for Science & Mutation Fitness Predictor & \underline{0.509953} & \textbf{0.518636} \\
        Time Series \& Forecasting & Exogenous-Variable Target Forecasting Model & \textbf{0.445408} & \underline{0.427787} \\
        Structured \& Causal Reasoning & Discrete Causal Graph Discovery & \underline{0.000000} & \textbf{0.371316} \\
        Trustworthy Learning & Training Regularization for Membership Privacy & \underline{0.360039} & \textbf{0.391151} \\
        \midrule
        Average & & \underline{0.321543} & \textbf{0.425198} \\
        \bottomrule
    \end{tabularx}
    \par\smallskip
    \begin{minipage}{\linewidth}
        \footnotesize
        Labels in Figure~\ref{fig:mls_transfer}: LM, Language Models; Classical, Classical and Adaptive Learning; RL, Reinforcement Learning; Opt., Optimization and Theory; Robotics, Robotics; Vision, Vision and Generation; DL, Deep Learning; Systems, Machine Learning Systems and Efficient Machine Learning; AI4Sci, Artificial Intelligence for Science; TS, Time Series and Forecasting; Causal, Structured and Causal Reasoning; Trust., Trustworthy Learning.
    \end{minipage}
    \endgroup
\end{table}

\noindent\textbf{Variation across optimization tasks.}
Table~\ref{tab:mls_full_results} shows that the largest absolute reward gains occur in causal discovery, language-model demasking, and multi-objective optimization.
Other tasks show smaller improvements, including behavioral cloning, activation-function design, and protein mutation effect prediction.
Both models receive zero reward on Gaussian splatting densification, while \sname{} scores lower on time-series forecasting.
The aggregate improvement thus reflects gains across several algorithmic components, with task-dependent benefits under the shared search budget.

\end{document}